\documentclass{bmvc2k}

\title{Global Aggregation then Local Distribution in Fully Convolutional Networks}

\addauthor{Xiangtai Li}{lxtpku@pku.edu.cn}{1}
\addauthor{Li Zhang}{lz@robots.ox.ac.uk}{2}
\addauthor{Ansheng You}{youansheng@pku.edu.cn}{1}
\addauthor{Maoke Yang}{maokeyang@deepmotion.ai}{3}
\addauthor{Kuiyuan Yang}{kuiyuanyang@deepmotion.ai}{3}
\addauthor{Yunhai Tong}{yhtong@pku.edu.cn}{1}
\usepackage{amssymb}
\usepackage{amsmath}
\addinstitution{
Key Laboratory of Machine Perception, \\
School of EECS,\\
Peking University
}
\addinstitution{
Department of Engineering Science,\\
Torr Vision Group,\\
University of Oxford
}
\addinstitution{
DeepMotion AI Research
}

\runninghead{Li \etal}{Global Aggregation then Local Distribution}

\def\eg{\emph{e.g}\bmvaOneDot}

\def\etal{\emph{et al}\bmvaOneDot}

\begin{document}

\maketitle
% The main argument for this paper is to say, adding a local module after the global module can be beneficial for many vision tasks such as semantic segmentation, object detection and instance segmentation.
\begin{abstract}
\noindent 
It has been widely proven that modelling long-range dependencies in fully convolutional networks (FCNs) via global aggregation modules is critical for complex scene understanding tasks such as semantic segmentation and object detection. However, global aggregation is often dominated by features of large patterns and tends to oversmooth regions that contain small patterns (\eg, boundaries and small objects). 
To resolve this problem, we propose to first use \emph{Global Aggregation} and then \emph{Local Distribution}, which is called GALD, where long-range dependencies are more confidently used inside large pattern regions and vice versa. 
The size of each pattern at each position is estimated in the network as a per-channel mask map. 
GALD is end-to-end trainable and can be easily plugged into existing FCNs with various global aggregation modules for a wide range of vision tasks, and consistently improves the performance of state-of-the-art object detection and instance segmentation approaches. In particular, GALD used in semantic segmentation achieves new state-of-the-art performance on Cityscapes test set with mIoU 83.3\%. 
Code is available at: \url{https://github.com/lxtGH/GALD-Net}

\end{abstract}

\section{Introduction}
Detection and segmentation tasks have made steady progress with more powerful representations learned from Fully Convolutional Networks (FCNs). Since stacking more convolutional layers is not an effective way to achieve large receptive fields for long-range dependency modeling~\cite{zhou2014object,luo2016understanding}, several Global Aggregation (GA) modules have  been proposed to resolve this problem.

In contrast to a standard convolutional layer which aggregates features in a small local window, GA modules use long-range operators such as averaging pooling~\cite{pspnet, deeplabv3} and spatial-wise feature propagation over the whole image~\cite{Nonlocal,ocnet,DAnet}. FCNs coupled with GA modules have consistently improved basic FCNs especially for large objects. 

\begin{figure*}
\centering
\includegraphics[width=0.99\linewidth]{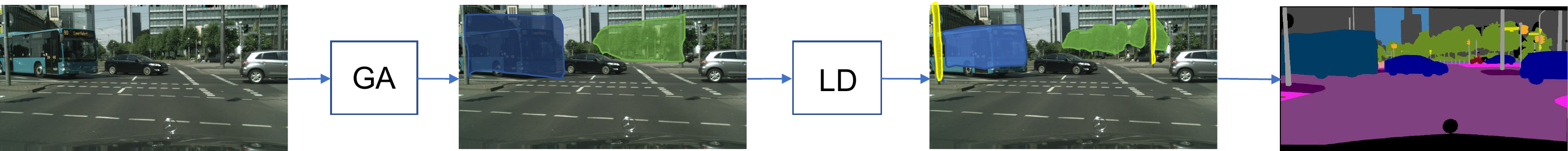}
\caption{Our proposed GALD framework for semantic segmentation task. The imbalanced spread of information from small and large patterns in GA module is appropriately handled through LD module.
}
\label{fig:teaser}
\end{figure*}

Unfortunately, the advantage of GA modules for large objects is a disadvantage for small patterns such as object boundaries and small objects, where features from GA modules tends to oversmooth the predictions for these small patterns. Thus, a straightforwards idea is using GA features conditionally on the pattern size of each position. Accordingly, we propose a Local Distribution (LD) module after a GA module (together as GALD for short) to adaptively distribute GA features at each position as illustrated in Fig.~\ref{fig:teaser}. The adaptive process is controlled by a set of mask maps, where each mask map is estimated from a feature map that records activations of some latent pattern over the whole image. 

LD is a simple and universal module, and can be combined with existing GA modules to form different GALD modules for various detection and segmentation tasks. In our experiment, LD is verified on GA modules such as PSP~\cite{pspnet}, ASPP~\cite{deeplabv2}, Non-Local~\cite{Nonlocal} and CGNL~\cite{cgnl}, and achieves consistent performance improvement. We also extensively verify GALD on three vision benchmarks, including Cityscapes for semantic segmentation, Pascal VOC 2007 for object detection, and MS COCO for both object detection and instance segmentation, and all achieve notable improvement. In particular, for semantic segmentation evaluated on Cityscapes test set, GALD achieves mIoU of $83.3\%$ with single model and ResNet101 as our backbone network, which surpasses all previously best published single-model results using ResNet101 as backbone network.

\section{Related Work}
\label{sec:related}
To keep spatial information required by detection and segmentation tasks, convolutional networks designed for image classification are modified to FCNs by removing global information aggregation layers such as global average pooling layer and fully-connected layers~\cite{fcn}. To quickly increase receptive field size while keeping the spatial resolution, filters in top convolutional layers are enlarged by dilation~\cite{deeplabv1, deeplabv2}. 

To further enlarge the receptive field to the whole image, several methods are proposed recently. Global average pooled features are concatenated into existing feature maps in ~\cite{parsenet}. In PSPnet~\cite{pspnet}, average pooled features of multiple window sizes including global average pooling are upsampled to the same size and concatenated together to enrich global information. The DeepLab series of papers \cite{deeplabv1, deeplabv2, deeplabv3} propose atrous or dilated convolutions and atrous spatial pyramid pooling (ASPP) to increase the effective receptive field. DenseASPP~\cite{denseaspp} improves on \cite{deeplabv2} by densely connecting convolutional layers with different dilation rates to further increase the receptive field of network. In addition to concatenating global information into feature maps, multiplying global information into feature maps also shows better performance~\cite{encodingnet, cbam, cgnl, dfn}.In particular, EncNet \cite{encodingnet} and DFN \cite{dfn} use attention along the channel dimension of the convolutional feature map to account for global context such as the co-occurrences of different classes in the scene. CBAM\cite{cbam} explores channel and spatial attention in cascade way to learn task specific representation.

Recently, advanced global information modeling approaches initiated from non-local network~\cite{Nonlocal} are showing promising results on scene understanding tasks. In contrast to convolutional operator where the information is aggregated locally defined by filters, the non-local operator aggregates information from the whole image based on an affinity matrix calculated among all positions around the image. Using non-local operator, impressive results are achieved in OCNet~\cite{ocnet},CoCurNet~\cite{CoCurrentNet}, DANet~\cite{DAnet}, A2Net~\cite{a2net}, CCnet~\cite{ccnet} and Compact Generalized Non-Local Net~\cite{cgnl}. OCNet~\cite{ocnet} uses non-local bolocks to learn pixel-wise relationship while CoCurNet~\cite{CoCurrentNet} adds extra global average pooling path to learn whole scene statistic. DANet~\cite{DAnet} explores orthogonal relationships in both channel and spatial dimension using non-local operator. CCnet~\cite{ccnet} models the long range dependencies by considering its surrounding pixels on the criss-cross path through a recurrent way to save both computation and memory cost. Compact Generalized non-local Net~\cite{cgnl} considers channel information into affinity matrix. Another similar work to model the pixel-wised relationship is PSANet~\cite{psanet}. It captures pixel-to-pixel relations using an attention module that takes the relative location of each pixel into account. 

Another way to get global representation is using graph convolutional networks, and do reasoning in a non-euclidean space~\cite{zhang2019dynamic,zhangli_dgcn,beyond_grids,graph_reason} where messages are passing between each node before projection back to each position. 
Glore~\cite{graph_reason} projects the feature map into interaction space using learned projection matrix and does graph convolution on projected fully connected graph. BeyondGrids~\cite{beyond_grids} learns to cluster different graph nodes and does graph convolution in parallel. DGCNet~\cite{zhangli_dgcn} proposes to use graph convolution network in both channel and spatial space to harvest different global context information.

All previous work focus on global context modeling, our work also utilizes global information modeling but takes a further step to better distribute the global information to each position, and further improves GA modules on both detection and segmentation tasks.
\section{Method}

\subsection{Model Overview}
Our method, Global Aggregation (GA) then Local Distribution (LD), dubbed GALD, exploits the long-range contextual information of the feature $ \mathbf{F} \in \mathbb{R}^{ H \times W \times C }$  from a fully-convolution network (FCN), and then adaptively distributed the global context to each spatial and channel position of the output feature, $\mathbf{F}_{GALD} \in \mathbb{R}^{ H \times W \times C }$.
To be noted, one can choose any one of the methods discussed in Section~\ref{sec:related} as GA.
% In this section, we mainly discuss the proposed LD module in details.

\begin{figure*}
\centering
\includegraphics[width=0.99\linewidth]{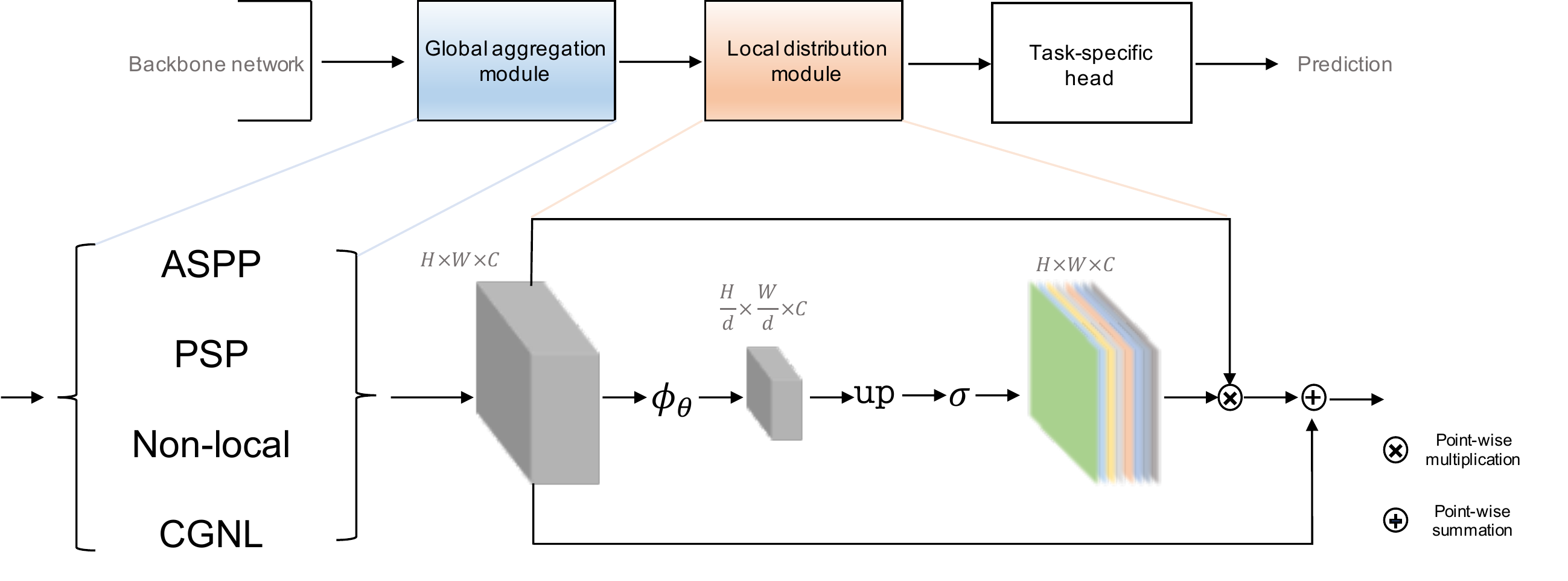}
\caption{Schematic illustration of GALD, which contains two main components: Global Aggregation (GA) and Local Distribution (LD). 
GALD receives a feature map from the backbone network and outputs a feature map with same size with global information appropriately assigned to each local position.
}
\label{fig:whole}
\end{figure*}

\subsection{GALD}
\noindent{\textbf{Global Aggregation.}} 
To calculate a feature vector for each position, GA module takes feature vectors of $\mathbf{F}$ in a large window even the whole feature map depending on different GA designs. Take the Compact Generalized Non-Local (CGNL)~\cite{cgnl} as an example, similar to non-local~\cite{Nonlocal}, it aggregates contextual information from all spatial and channel positions in the same group. 
Specifically,  the global statistics are calculated for each group and multiplied back to the features in the same group, which forms $\mathbf{F}_{GA}$. In our implementation, we downsample $\mathbf{F}$ by a factor of $2$ for saving memory and computation cost without observing performance degradation, which also demonstrates the coarse property of global aggregation. 

Since GA modules calculate global statistics of features in large windows, they are easily biased towards features from large patterns as they contain more samples. Then the global information distributed to each position is also biased towards large patterns, which causes over-smoothing results for small patterns. One can refer to Section~\ref{sec:ablation} for more detailed visualization results.

\noindent{\textbf{Local Distribution.}} 
LD is proposed to adaptively use $\mathbf{F}_{GA}$ considering patterns on each position. Without explicit supervision, the required patterns are latently described by $C$ channels in $\mathbf{F}_{GA}$. For each pattern $c\in\{1,...,C\}$, a spatial operator is learned to recalculate the spatial extent of the pattern in an image based on the activation map  $\mathbf{F}_{GA}[:,:,c]$ sliced from $\mathbf{F}_{GA}$. Intuitively, spatial operators for large patterns would shrink the spatial extent more while shrinking less even expand for small patterns.

The spatial operators for each pattern/channel is modeled as a set of depth-wise convolutional layers with $\mathbf{F}_{GA}$ as input, i.e.,
\begin{align}
    \mathbf{M} = \sigma(\text{upsample}(\mathbf W_{d} \mathbf{F}_{GA})),
\end{align}
where $\mathbf{M} \in [0,1]^{H\times W \times C}$ contains the mask maps for each pattern and describes the recalculated spatial extents of each pattern, $\sigma(\cdot)$ is the sigmoid function, $\mathbf W_{d}$ is the weights of  those depth-wised convolutional filters with $d$ as the downsampling rate by stride convolution. The output mask $\mathbf{M}$ is sensitive to both spatial and channel and it is upsampled using bilinear interpolation. With the mask maps $\mathbf{M}$, $\mathbf{F}_{GA}$ is refined into $\mathbf{F}_{GALD}$ by
\begin{align}
    \mathbf{F}_{GALD} = \mathbf{M} \odot \mathbf{F}_{GA}  + \mathbf{F}_{GA},
\end{align}
where $\odot$ the element-wise multiplication, and elements in $\mathbf{F}_{GA}$ are weighted according the estimated spatial extent of each pattern at each position. In summary, LD predicts local weights $\mathbf{M}$ for each position of GA features
and avoids issues of coarse feature representation. 

As a common practice~\cite{pspnet}, original feature $\mathbf{F}$ and global aggregated feature $\mathbf{F}_{GA}$ are concatenated together for final task-specific head, i.e.,
\begin{align}
\begin{split}
    \mathbf{F}_{o} &= \text{concat}( \mathbf{F}_{GALD}, \mathbf{F}) \\
                   &= \text{concat}( \mathbf{M} \odot \mathbf{F}_{GA}  + \mathbf{F}_{GA},\mathbf{F} ),
\end{split}
\label{eq:final_feature}
\end{align}
where $\mathbf{M}$ adds point-wise trade-off between global information $\mathbf{F}_{GA}$ and local detailed information $\mathbf{F}$. Note that since the lack of details in GA, LD module only changes the proportion and distribution of coarse features in GA and leads to a fine-grained feature representation output $\mathbf{F}_{o}$.

\subsection{Overall Architecture}
Fig.~\ref{fig:whole} illustrates the overall architecture with GALD. For semantic segmentation, GALD is added right after a FCN, features from Eq.~\ref{eq:final_feature} are used for final prediction. To further boost the performance, Online Hard Example Mining (OHEM) loss~\cite{SegOHEM} is used for training, where only top-K ranked pixels according their losses are used during back-propagation.

For object detection and instance segmentation task, GALD is added at the end of stage4 of a ResNet backbone, FPN~\cite{fpn} is used to build a strong baseline with a feature pyramid for multi-scale object detection. $\mathbf{F}_{GALD}$ sits on top of FPN and passes information from the top-down pathway.

\section{Experiment}
In this section, we verify GALD on three scene understanding tasks including semantic segmentation, object detection and instance segmentation.

\subsection{Benchmarks} 
\textbf{Cityscapes:} Cityscapes~\cite{Cityscapes} is a benchmark that densely annotated for 19 categories in urban scenes, which contains 5000 fine annotated images in total and is divided into 2975, 500, and 1525 images for training, validation and testing, respectively. In addition, 20,000 coarse labeled images are also provided to enrich the training data. Images of this dataset are all with the same high resolution, i.e., $1024 \times 2048$. Following the standard protocol~\cite{Cityscapes}, mean Intersection over Union (mIoU) of all categories on validation set and test set is used for performance comparison. 

\noindent{\textbf{MS COCO:}} MS COCO~\cite{COCO_dataset} is built for detecting and segmenting objects found in everyday life in their natural environment. The dataset for detection consists of three sets for 80 common object categories, i.e., the training set has 118,287 images, validation set has 5,000 images and test-dev set has more than 20,000 images.

\noindent{\textbf{Pascal VOC:}} Pascal VOC~\cite{VOC} is a widely used public benchmark for semantic segmentation and object detection covering $20$ object categories including the background. We use VOC 2007 and VOC 2012 trainval set as training set and report results on VOC 2007 test set. 

\begin{figure*}
\centering
\includegraphics[width=0.99\linewidth]{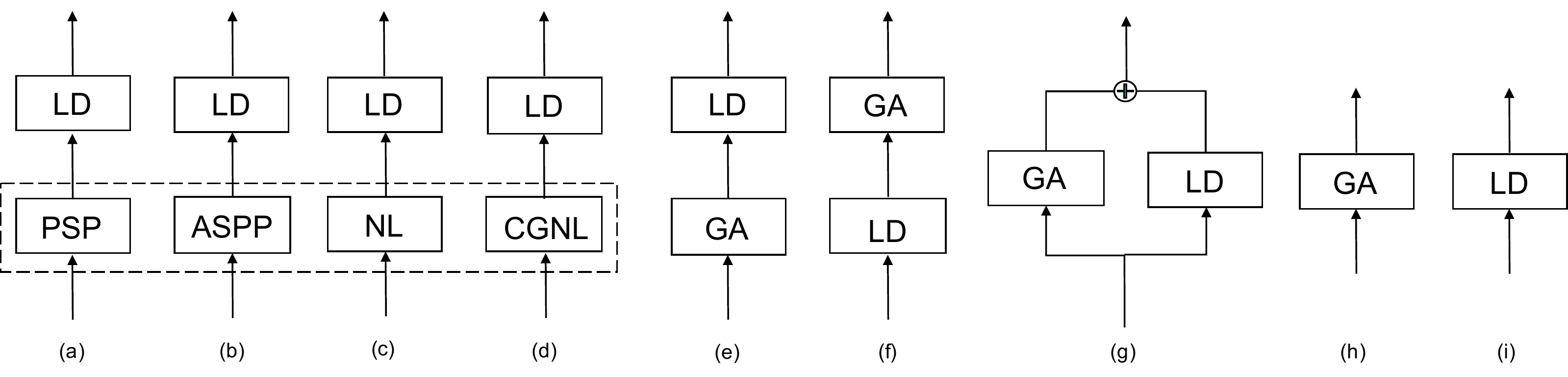}
\caption{Ablation studies on combinations of GA and LD. (a)-(d) shows the different GA modules with LD. (e)-(g) shows the different arrangements of GA and LD. (h)-(f) represents using GA and LD respectively.
}
\label{fig:ablation}
\end{figure*}

\subsection{Implementation Details}

{\bfseries Semantic Segmentation } We employ Fully Convolutional Networks (FCNs) as baseline, where ResNet pretrained on ImageNet is chosen as the backbone following the same setting as PSPNet~\cite{pspnet}, the proposed GALD is appended to the backbone with random initialization. For optimization, we also keep the same setting as PSPNet, where mini-batch SGD with momentum 0.9 and initial learning rate 0.01 is used to train all models with 50K iterations, using mini-batch size of 8 and crop size of 769. During training, ``poly'' learning rate scheduling policy where power = 0.9 is used to adjust the learning rate. Synchronized batch normalization~\cite{encodingnet} is used for better mean / variance estimation across GPUs.

\noindent{{\bfseries Object Detection and Instance Segmentation}} For object detection and instance segmentation, mmdetection~\cite{mmdetection2018} is used as our baseline implementation for fair comparison. GALD is evaluated for object detection on Pascal VOC based on Faster R-CNN, and for both object detection and instance segmentation on MS COCO based on Mask R-CNN. FPN~\cite{fpn} is used as default setting in all these experiments. For fair comparison, we report all the results that we re-implemented in our framework.

\begin{table}[!t]
	\centering			
	\begin{minipage}{\dimexpr.40\linewidth}
		\centering
		\tiny
		\resizebox{1.0\textwidth}{!}{%
			\begin{tabular}{ l|c|c }
				\hline
				Method & mIoU(\%) & $\Delta a$ \\
				\hline
				FCN (Baseline) & 73.7  & - \\
				\hline \hline
				+ASPP~\cite{deeplabv3} & 77.2 & 3.5 $\uparrow$ \\ 
				
				+NL~\cite{Nonlocal} & 78.0 & 4.3 $\uparrow$ \\
				%\hline
				+PSP~\cite{pspnet} &  76.2 & 2.5 $\uparrow$ \\
				%\hline
				+CGNL~\cite{cgnl} & \textbf{78.2} & \textbf{4.5} $\uparrow$ \\
				\hline
			\end{tabular}
		}
		\par
		{\footnotesize(a) Ablation study on different GA modules using ResNet50 as backbone.}
	\end{minipage}
	\begin{minipage}{\dimexpr.52 \linewidth}
		\centering
		\tiny
		\resizebox{1.0\textwidth}{!}{%
			\begin{tabular}{ l|c|c|c }
				\hline
				Method & mIoU(\%) & $\Delta a$ & $\Delta b$\\
				\hline
				FCN (Baseline) & 73.7  & - & -\\
				\hline
				\hline
				+LD & 77.5 & 3.8 $\uparrow$ & - \\
				+PSP + LD & 78.9 & 5.2 $\uparrow$ & \textbf{2.7} $\uparrow$ \\ 
				+ASPP + LD & 79.5 & 5.4 $\uparrow$ & 2.3 $\uparrow$ \\
				+NL + LD & 79.2 & 5.3 $\uparrow$ & 1.2 $\uparrow$ \\
				%\hline
				+CGNL + LD & \textbf{79.6} & \textbf{5.9 $\uparrow$} & 1.4 $\uparrow$ \\
				%(f) Baseline + 2 GALD & 78.7 & 5.0 $\uparrow$ \\
				\hline
			\end{tabular}
		}\par
		{\footnotesize(b) Ablation study on LD applied on different GA modules using ResNet50 as backbone.}
		
	\end{minipage}
	\vspace{3mm}
	
	\begin{minipage}{\dimexpr.40 \linewidth}
		\centering
		\tiny
		\resizebox{1.0\textwidth}{!}{%
			\begin{tabular}{ l|c|c}
				\hline
				Method & mIoU(\%) & $\Delta a$\\
				\hline
				FCN (Baseline) & 73.7  & - \\
				\hline
				\hline
				+Parallel & 77.5 & 3.8 $\uparrow$\\ 
				+LDGA  & 78.1 & 4.4 $\uparrow$ \\
				+GALD  & \textbf{79.6} & \textbf{5.9 $\uparrow$}\\
				\hline
			\end{tabular}
		}\par
		{\footnotesize(c) Ablation study on different arrangements of GA and LD using ResNet50 as backbone.}
		
	\end{minipage}	
	\begin{minipage}{\dimexpr.50 \linewidth}
		\centering
		\tiny
		\resizebox{1.0\textwidth}{!}{%
			\begin{tabular}{ l|c|c|c}
				\hline
				Method & mIoU(\%) & Backbone & $\Delta a$\\
				\hline
				FCN (Baseline) & 73.7 & ResNet50  & -\\
				FCN (Baseline) & 75.3 & ResNet101 & - \\
				\hline
				\hline
				+CGNL & \textbf{79.7} & ResNet101 &  4.4 $\uparrow$  \\
				+CGNL+LD  & \textbf{79.6} & ResNet50 & \textbf{5.9 $\uparrow$}\\
				+PSP & 78.6 & ResNet101 & 4.9 $\uparrow$ \\
				+PSP+LD & 78.9 & ResNet50 & 5.2 $\uparrow$ \\
				\hline
			\end{tabular}
		}\par
		{\footnotesize(d) Ablation study on different backbones.}
		
	\end{minipage}
	\vspace{3mm}
	
	\begin{minipage}{\dimexpr.50 \linewidth}
		\centering
		\tiny
		\resizebox{1.0\textwidth}{!}{%
			\begin{tabular}{ l|c|c}
				\hline
				Method & mIoU(\%)  & $\Delta b$\\
				\hline
				FCN (Baseline) & 73.7  & -\\
				FCN + CGNL & 78.2 & - \\
				\hline
				\hline 
				+CGNL+LD(depth-wise convolution)  & \textbf{79.6} & \textbf{1.4 $\uparrow$}\\
				+CGNL+LD(bilinear interpolation) & 77.6  & 0.6 $\downarrow$ \\
				+CGNL+LD(average pooling) & 76.5  & 1.7 $\downarrow$ \\
				\hline
			\end{tabular}
		}\par
		{\footnotesize(e) Ablation study on downsampling strategies for mask estimation in LD using ResNet50 as backbone, where the downsamping ratio is 8.}
	\end{minipage}
	\vspace{3mm}
	
	\caption{Comparison results on Cityscapes validation set, where $\Delta a$ denotes the performance difference comparing with baseline, and $\Delta b$ denotes performance difference between using GALD module and the corresponding GA module.
		All methods are evaluated with single-scale crop test.}
	\vspace{-5mm}
	\label{tab:city_ablation}
\end{table}

\subsection{Results on Cityscapes}
Two groups of experiments are conducted on Cityscapes, the first group of experiments verifies the effectiveness of our GALD framework by ablation studies.  The second group of experiments compares GALD to the state-of-the-art methods.

\subsubsection{Ablation Studies }
\label{sec:ablation}
\textbf{Comparison with baseline} 
We explore our LD module with four different GA modules as illustrated in Fig.~\ref{fig:ablation} (a)-(d).
Table~\ref{tab:city_ablation}(a) first reports the performances of adding four GA modules to the baseline FCN, where all methods are using the same backbone ResNet50 for fair comparison.  Obviously, all GA modules significantly improves the baseline FCN on semantic segmentation task, where CGNL performs better than other three GA modules.
Table~\ref{tab:city_ablation}(b) reports the results by adding our proposed LD module. Directly using LD alone improves the baseline FCN by 3.8\%, which demonstrates that features from FCN have the similar problem as features from GA modules. LD together with four different GA modules consistently improves the corresponding GA module. Comparing with baseline, the combination of CGNL+LD achieves the best performance, and we mainly choose CGNL as our GA module in following experiments.

\begin{figure*}
	\centering
	\includegraphics[width=1.0\linewidth]{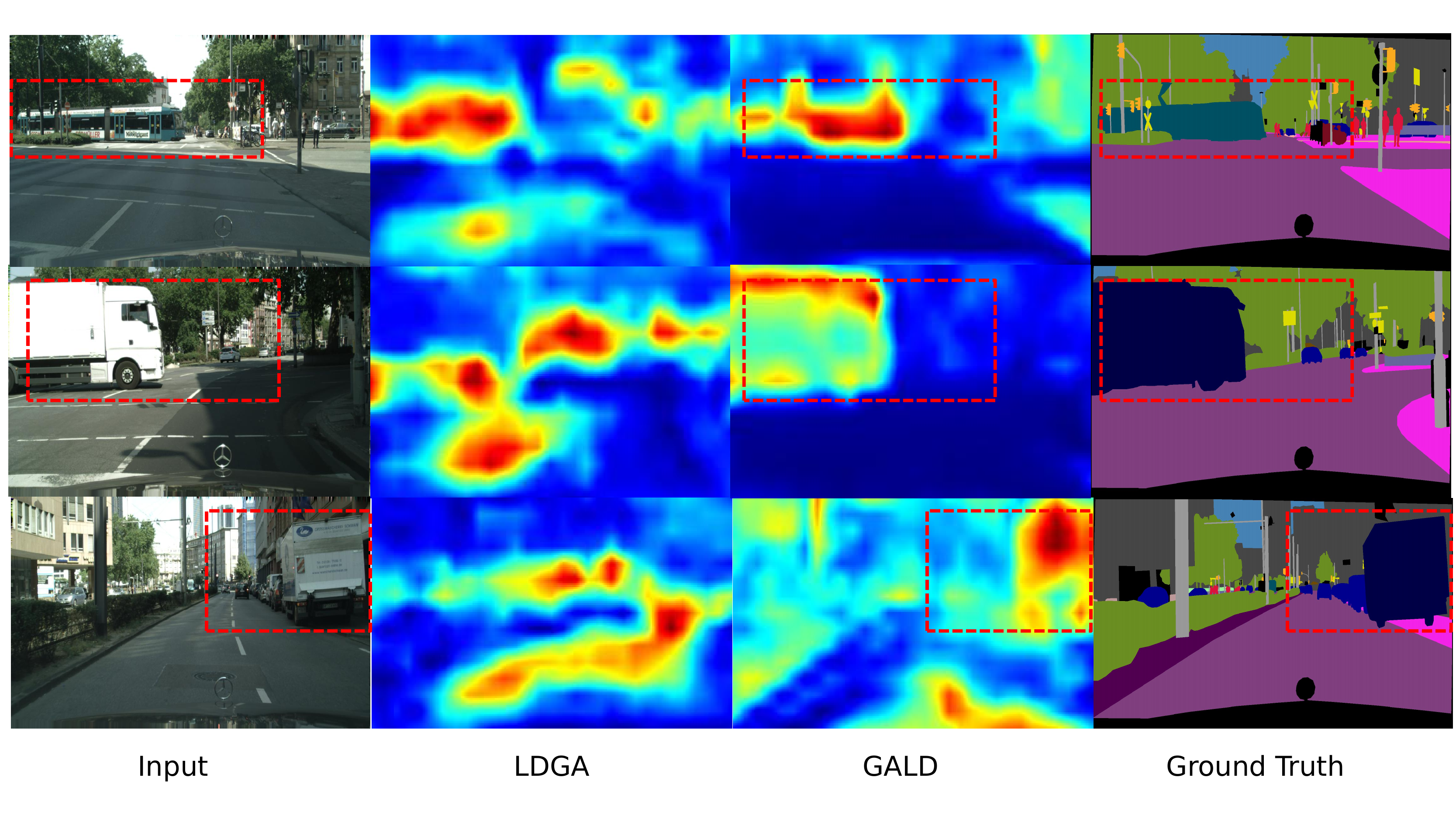}
	\caption{
	    Comparison of mask maps learned in different arrangements of GA and LD. The mask maps are calculated by the mean of $\textbf{M}$ along channel dimension.
		Best view in color.}
	\label{fig:attenion_mask_2}
\end{figure*}

\noindent{{\bfseries Arrangements of LD and GA}} Considering LD module can also improve the baseline, we further study different arrangements of LD and GA as illustrated in Fig.~\ref{fig:ablation} (e)-(g). (f) and (g) represent LDGA and Parallel in Table.~\ref{tab:city_ablation}(c) respectively. LDGA means first doing LD then doing GA while Parallel concatenates the output of LD and GA.
Table.~\ref{tab:city_ablation}(c) reports the results of the three different arrangements, where all improve the baseline and GALD achieves best result. Fig.~\ref{fig:attenion_mask_2} shows the mask maps learned in LDGA and GALD, where mask maps
learned by GALD are more focused on regions inside large objects then weight global features more in these regions, while mask maps from LDGA have no obvious focus on large objects since the LD module has not accessed to global feature yet.

\noindent{\bfseries Compared with stronger backbone} To further prove the effectiveness of our method, we compare GALD using ResNet50 as backbone with a stronger backbone ResNet101 in Table~\ref{tab:city_ablation}(d). Our method achieves similar performance improvement comparing GA modules with stronger backbone which further prove the effectiveness of LD module.

\noindent{\bfseries Comparison with different downsampling strategies} We also explore three different downsampling strategies for LD, including average pooling, bilinear interpolation and depth-wise stride convolution. Table~\ref{tab:city_ablation}(e) reports the comparison results, depth-wise stride convolution achieves the best result, while average pooling and bilinear interpolation even slightly degrades the performance, which shows that the learnable filters for each channel is important to refine the features from the GA module.

\begin{figure*}
	\centering
	\includegraphics[width=1.0\linewidth]{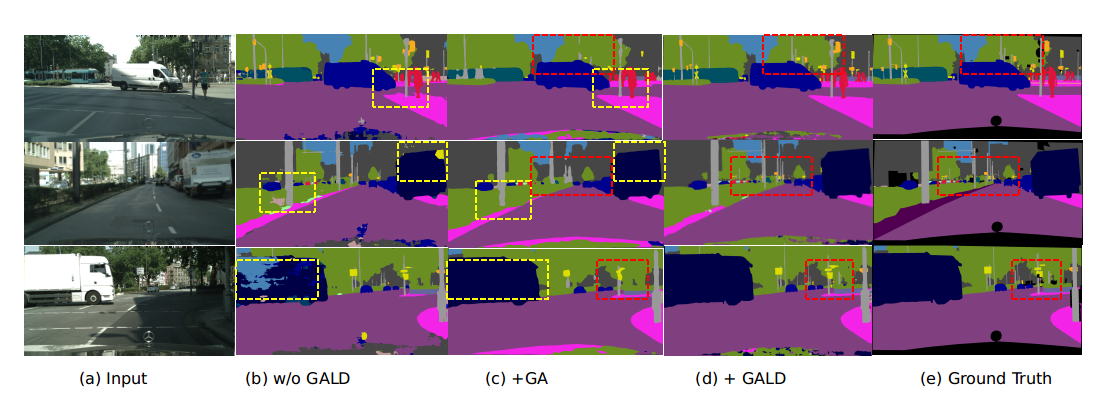}
	\caption{
		Visualization of different parts output results in one model.(a), input images; (b),results after FCN's outputs; (c), results after GA module's outputs; (d), results after GALD module'ss outputs;
		(e), ground truth. Yellow boxes highlight regions that GA can handle global semantic consistency, while red boxes highlight regions that LD can recover more detailed information.
		Best view in color.}
	\label{fig:attenion_mask}
\end{figure*}

\noindent{\bfseries Visualization of GALD}
To further study the features at different stages, we add another two segmentation heads on features outputted from FCN and GA respectively, the model is fine tuned until converge to analyze segmentation ability of features from different stages. Figure~\ref{fig:attenion_mask} compares the segmentation results, segmentation based on GA resolves the ambiguities in FCN features but also tends to over smoothing regions of small patterns which are shown in red boxes. Segmentation of GALD keeps the global structure of GA while refines back the details.

\subsubsection{Comparison with state-of-the-art}

\begin{table}
	\centering
	\begin{minipage}{\dimexpr.40\linewidth}
		\centering
		\tiny
		\resizebox{1.0\textwidth}{!}{%
				\begin{tabular}{l|c|c}
					\hline
					Method & Backbone & mIoU(\%)  \\
					\hline
					SAC~\cite{sac}\textdagger & ResNet101 &  78.1 \\ 
					AAF~\cite{aaf}\textdagger  & ResNet101 &  79.1 \\ 
					BiSeNet~\cite{bisenet}\textdagger & ResNet101 &  78.9 \\ 
					PSANet~\cite{psanet}\textdagger & ResNet101 &  80.1 \\ 
					DFN~\cite{dfn}\textdagger & ResNet101 &  79.3 \\ 
					DenseASPP~\cite{denseaspp}\textdagger & DenseNet161 & 80.6 \\
					Glore~\cite{graph_reason}\textdagger & ResNe50 & 79.5 \\
					Glore~\cite{graph_reason}\textdagger & ResNet101 & 80.9 \\
					DAnet~\cite{DAnet}\textdagger & ResNet101 & 81.5 \\
					\hline
					GALDNet\textdagger & ResNet50 & \textbf{80.8}  \\ 
					GALDNet\textdagger & ResNet101 & \textbf{81.8} \\
					\hline
				\end{tabular}
		}
		\par
		{\footnotesize(a) Results on Cityscapes test server trained with fine-data.}
	\end{minipage}
	\begin{minipage}{\dimexpr.52 \linewidth}
		\centering
		\tiny
		\resizebox{1.0\textwidth}{!}{%
			\begin{tabular}{l|c|c}
				\hline
				Method & Backbone & mIoU(\%)  \\
				\hline
				PSP~\cite{pspnet}\textdaggerdbl & ResNet101 &  81.2 \\ 
				Deeplabv3+~\cite{deeplabv3p}\textdaggerdbl & Xception & 82.1 \\
				DPC~\cite{DPC}\textdaggerdbl & Xception & 82.6 \\
				Auto-Deeplab~\cite{auto-deeplab}\textdaggerdbl & - & 82.1 \\
				%DRN~\cite{DRN}\textdaggerdbl & WideResNet38 & 82.8 \\
				\hline\hline
				GALDNet\textdaggerdbl & ResNet101 & \textbf{82.9} \\
				\hline
				GALDNet(+Mapillary)\textdaggerdbl & ResNet101 & \textbf{83.3} \\
				\hline
			\end{tabular}
		}\par
		{\footnotesize(b) Results on Cityscapes test server trained with both fine and coarse data}
		
	\end{minipage}
	\vspace{3mm}
	\caption{State-of-the-art comparison experiments on Cityscapes test set. \textdagger means training with only the train-fine dataset. \textdaggerdbl means training with both the train-fine and coarse data}
	\label{tab:cityscapes_results}
\end{table}

We further compare our results with other state-of-the-art methods in this section. We choose dilated ResNet50 and ResNet101 as backbone models. The results are summarized in Table~\ref{tab:cityscapes_results}. For fair comparison, we first compare methods trained with only fine annotation data in Table~\ref{tab:cityscapes_results}(a), and then compare the results with other methods using extra training data in Table~\ref{tab:cityscapes_results}(b). Following~\cite{pspnet}, multi-scale crop test is used for final test submission.
As illustrated, our method surpasses all previous methods. In particular, our model based on a weak backbone ResNet50 can still achieve comparable performance, which is higher than most methods with stronger backbone. By using extra coarse annotation data for training, our method achieves 82.9\% mIoU, which also surpasses the state-of-the-art methods. By further adding Mapillary~\cite{mapillary} as training data, the proposed method achieves 83.3\% mIoU based on ResNet101. To the best of our knowledge, this is the first single model using ResNet101 as backbone that surpasses 83\% mIoU on Cityscapes test server. More detailed per-class results, visualization results and training settings can be referred to the supplementary material.

\subsection{Results on Pascal VOC and COCO dataset}
 { \bfseries Pascal VOC:} 
 We perform experiments on the PASCAL VOC 2007 data set to evaluate the effect of GALD for object detection. We train all the models on the union set of VOC 2007 trainval and VOC 2012 trainval (07+12) for 14 epochs with weight decay of 0.0001 and momentum of 0.9. 
 For comparison, experiments of non-local block \cite{Nonlocal} are also summarized and are denoted as NL. 
 As results listed in Table~\ref{tab:voc_coco}(a), GALD consistently improves detection accuracy over the strong baseline Faster-RCNN using both ResNet50 and ResNet101 as backbone, which demonstrates the effectiveness of GALD for object detection. 

\noindent{ {\bfseries COCO:}} 
To further verify the generality of GALD, we conduct the experiments on instance segmentation task on MS COCO based on the state-of-the-art method Mask R-CNN. 
%Similar to the experiments on Pascal VOC, GALD is inserted between the backbone and FPN and the new model is trained with standard setting. 
Table~\ref{tab:voc_coco}(b) summarizes the AP of bounding box (AP-box) and AP of mask (AP-mask) evaluated on COCO minival. GALD improves the baseline by about 1\% regardless the used backbone. 
Figure~\ref{fig:seg_results}(b) compares the object detection and instance segmentation results of our method with baseline. With GALD, Mask R-CNN can find objects that are missed in baseline (e.g., the ``light'' in the third column), resolve ambiguity in region classification (e.g., the ``bed'' in the first column) and help to better estimate the spatial contents for objects (e.g., ``bear'' in last column).

\begin{table}
	\centering
	\begin{minipage}{\dimexpr.40\linewidth}
		\centering
		\tiny
		\resizebox{1.0\textwidth}{!}{%
			\begin{tabular}{l|c|c}
				\hline
				Backbone& Detector & mAP@.5  \\
				\hline
				ResNet50 & Faster-RCNN & 80.6   \\
				%\hline
				ResNet50 & + NL & 81.3  (0.7 $\uparrow$) \\
				%\hline
				ResNet50 & +CGNL & 81.1 (0.5 $\uparrow$) \\
				ResNet50 & + GALD  & \textbf{81.5}  (0.9 $\uparrow$) \\
				\hline \hline
				ResNet101 & Faster-RCNN & 80.7  \\
				%\hline
				ResNet101 & + NL & 82.3  (1.6 $\uparrow$)  \\
				%\hline
				ResNet101 & + GALD & \textbf{83.0}  (2.3 $\uparrow$) \\
				\hline
			\end{tabular}
		}
		\par
		{\footnotesize(a) Object detection results on VOC 2007 test set measured by mAP(\%),  Faster-RCNN with FPN serves as the baseline.}
	\end{minipage}
	\begin{minipage}{\dimexpr.50 \linewidth}
		\centering
		\tiny
		\resizebox{1.0\textwidth}{!}{%
			\begin{tabular}{l|c|c|c}
				\hline
				{Backbone} & {Detector} & {AP-box} & {AP-mask} \\
				\hline
				ResNet50 & Mask-RCNN & 38.2  & 34.8 \\
				% \hline
				ResNet50 & + NL & 39.0  (0.8 $\uparrow$)   & 35.3  (0.8 $\uparrow$)  \\
				ResNet50 & + CGNL & 38.9 (0.7 $\uparrow$) & 35.4 (0.6$\uparrow$) \\ 
				% \hline
				ResNet50 & + GALD  & \textbf{39.2}  (1.0 $\uparrow$)  & \textbf{35.6} (1.1 $\uparrow$)  \\
				\hline
				\hline
				ResNet101 & Mask-RCNN & 40.2 & 36.3 \\
				% \hline
				ResNet101 & + NL & 40.9   (0.7 $\uparrow$)  & 37.2   (0.9 $\uparrow$) \\
				% \hline
				ResNet101 & + GALD & \textbf{41.1}   (0.9 $\uparrow$) & \textbf{37.8} (1.5 $\uparrow$) \\
				\hline
			\end{tabular}
		}\par
		{\footnotesize(b) Results of object detection and instance segmentation on COCO dataset. Our method can improve Mask-RCNN baseline by around 1\%  across different backbones.}
		
	\end{minipage}
	\vspace{3mm}
	%	\captionsetup{font=footnotesize}
	\caption{Results on Pascal VOC dataset (a) and MS COCO dataset (b). }
	\label{tab:voc_coco}
\end{table}

\begin{figure}
	\centering
		\includegraphics[width=0.68\textwidth]{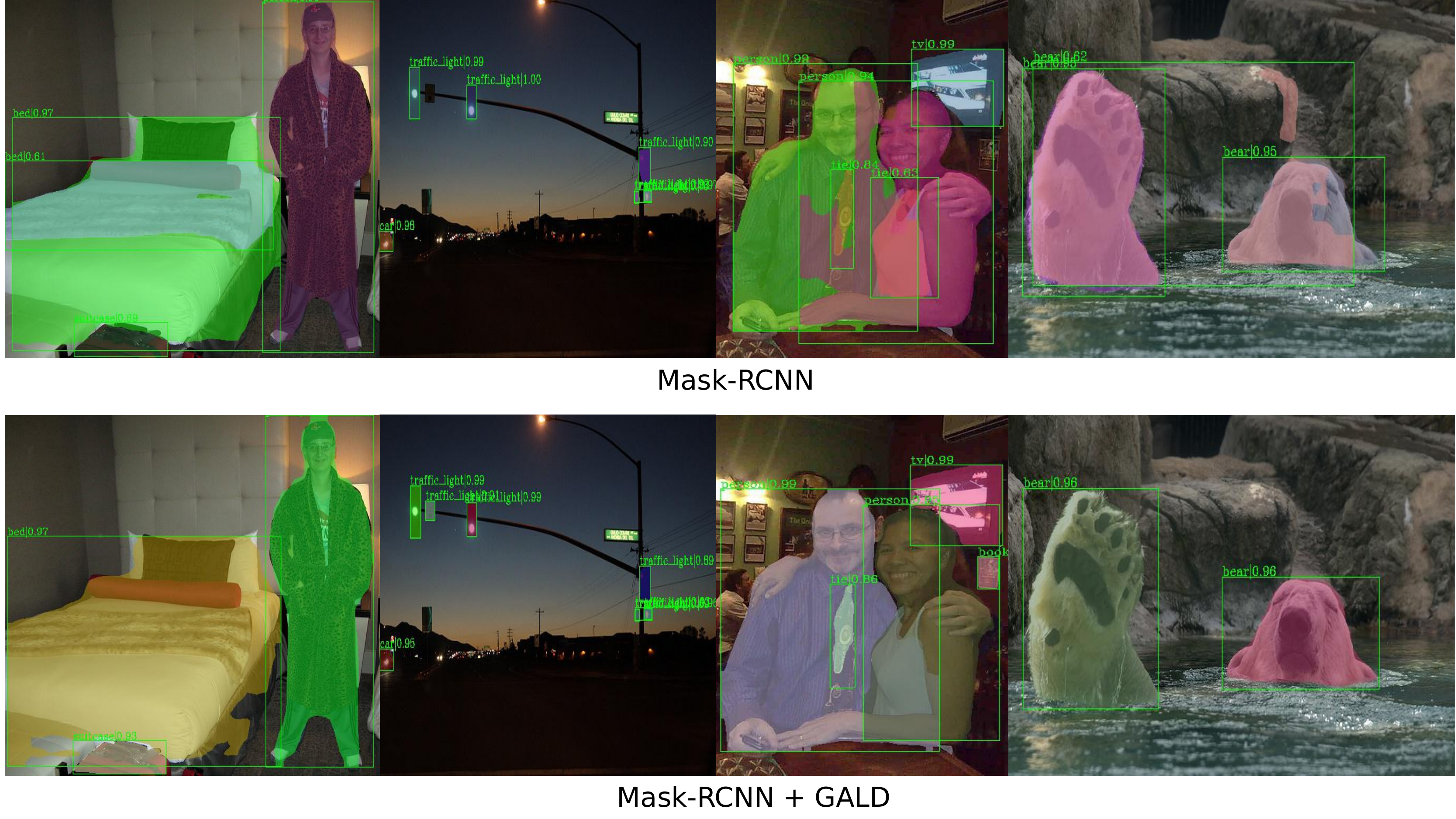}
	\caption{Comparison of object detection and instance segmentation results on MS COCO.}
	\label{fig:seg_results}
\end{figure}

\iffalse
\vspace{8mm}
\begin{figure}
	\centering
	\begin{tabular}{cc}
		\bmvaHangBox{\includegraphics[width=0.48\textwidth]{fig/seg_fcn_vis.png}}&
		\bmvaHangBox{\includegraphics[width=0.48\textwidth]{fig/coco.pdf}}\\
		(a) FCN with ResNet50 as backbone. & (b) Mask-RCNN with ResNet50 as backbone.
	\end{tabular}
	\caption{(a) Visualization of results on Cityscapes. (b) Visualization of results on MS COCO.}
	\label{fig:seg_results}
\end{figure}
\fi
\section{Conclusion}
In this paper, we propose GALD to adaptively distribute global information to each position for scene understanding tasks. In contrast to existing methods that assign global information uniformly to each position and cause the problem of blurring, GALD learns a set of mask maps to distribute global information adaptively according pattern distributions over the image. GALD benefits from both the GA module for ambiguity resolving and LD module for detail refinement. Extensive experiments verify the universality of GALD in improving the performance of semantic segmentation, object detection and instance segmentation. 
In the future, we will study the effectiveness of GALD for more vision tasks where both global and local information are important such as depth estimation.

\section*{Acknowledgments}
We gratefully acknowledge the support of DeepMotion AI Research for providing the computing resources in carrying out this research.
LZ is supported by EPSRC Programme Grant Seebibyte EP/M013774/1.

\section{Appendix}
\subsection{Detailed Results on Cityscapes}
\subsubsection{Training with Coarse labeled Data}
Ctiyscapes provides about 20000 coarse labeled images for training. To verify the 
both capacity and generality, we further fine tune our model on coarse data set.
Different from training on fine data set, we set batch size 16 and fix batch normalization layers in our model for about 50000 iterations using larger crop size. Then we fine tune the model back on the fine data set for 10000 iterations with lower learning rate. When we submit on test server, we use multi scale crop test with flip images input to get more accurate results. Finally we get better results and our single model can achieve 82.9\% mIoU.
\subsubsection{Training with Mapillary Data}
Mapillary~\cite{mapillary} is another city scene data containing 65 different labels. Here we only use 19 classes of 65 labels which are in cityscape category. Again we following the same steps in previous part,
we get more accurate model and our single model with ResNet101 as backbone can achieve 83.3\% mIoU ranked 3-rd in cityscapes leaderboard by the time of paper publication.
\subsubsection{Detailed Results}
Models trained with only fine-data set are shown in Table~\ref{tab:cityscapes_results_detail_fine}.
Models trained with extra data sets including COCO and Mapillary data sets are shown in Table~\ref{tab:cityscapes_results_detail_coarse}.

\begin{table*}[t]
	\scriptsize
	\centering
	\setlength{\tabcolsep}{0.9pt}
	\begin{tabular}{ l | c c c c c c c c c c c c c c c c c c c | c}
		\hline
		Method & road & swalk & build & wall & fence & pole & tlight & sign & veg. & terrain & sky & person & rider & car & truck & bus & train & mbike & bike & mIoU \\
		\hline
		ResNet38~\cite{resnet38} & 98.5 & 85.7 & 93.0 & 55.5 & 59.1 & 67.1 & 74.8 & 78.7 & 93.7 & 72.6 & 95.5 & 86.6 & 69.2 & 95.7 & 64.5 & 78.8 & 74.1 & 69.0 & 76.7 & 78.4 \\
		PSPNet~\cite{pspnet} & 98.6 & 86.2 & 92.9 & 50.8 & 58.8 & 64.0 & 75.6 & 79.0 & 93.4 & 72.3 & 95.4 & 86.5 & 71.3 & 95.9 & 68.2 & 79.5 & 73.8 & 69.5 & 77.2 & 78.4\\
		AAF~\cite{aaf} & 98.5 & 85.6 & 93.0 & 53.8 & 58.9 & 65.9 & 75.0 & 78.4 & 93.7 &
		72.4 & 95.6 & 86.4 & 70.5 & 95.9 & 73.9 & 82.7 & 76.9 & 68.7 & 76.4 & 79.1 \\
		SegModel~\cite{segmodel} & 98.6 & 86.4 & 92.8 & 52.4 & 59.7 & 59.6 & 72.5 & 78.3 & 93.3 & \textbf{72.8} & 95.5 & 85.4 & 70.1 & 95.6 & 75.4 & 84.1 & 75.1 & 68.7 & 75.0 & 78.5 \\
		DFN~\cite{dfn} & - & - & - & - & - & - & - & - & - & - & - & - & - & - & - & - & - & - & - & 79.3 \\
		BiSeNet~\cite{bisenet} & - & - & - & - & - & - & - & - & - & - & - & - & - & - & - & - & - & - & - & 78.9 \\
		PSANet~\cite{psanet} & - & - & - & - & - & - & - & - & - & - & - & - & - & - & - & - & - & - & - & 80.1 \\
		DenseASPP~\cite{denseaspp}  & 98.7 & 87.1 & 93.4 & 60.7 & 62.7 & 65.6 & 74.6 & 78.5 & 93.6 & 72.5 & 95.4 & 86.2 & 71.9 & 96.0 & \textbf{78.0} & 90.3 & 80.7 & 69.7 & 76.8 & 80.6 \\
		%final result 
		\hline
		Ours(ResNet50) & \textbf{98.7} & 86.8 & 93.4 & 57.6 & \textbf{63.1} & 68.7 & 76.1 & 80.3 & 93.6 & 72.3 & 95.4 & 87.0 & 72.2 & 96.1 & 75.4 &88.2 & 77.8 & 68.8 & 76.4 & \textbf{80.8} \\
		\hline
		Ours(ResNet101) & \textbf{98.7} & \textbf{87.2} & \textbf{93.8} & \textbf{59.3} & 61.9 & \textbf{71.4} & \textbf{79.2} & \textbf{82.0} & \textbf{93.9} & \textbf{72.8} & \textbf{95.6} & \textbf{88.4} & \textbf{74.8} & \textbf{96.3} & 74.1 & \textbf{90.6} & \textbf{81.1} & \textbf{73.4} & \textbf{79.8} & \textbf{81.8} \\
		\hline
	\end{tabular}
	\caption{Per-category results on Cityscapes test set. Note that all the models are trained with only fine-data .}
	\label{tab:cityscapes_results_detail_fine}
\end{table*}

\begin{table*}[t]
	\scriptsize
	\centering
	\setlength{\tabcolsep}{1.0pt}
	\begin{tabular}{ l | c c c c c c c c c c c c c c c c c c c | c}
		\hline
		Method & road & swalk & build & wall & fence & pole & tlight & sign & veg. & terrain & sky & person & rider & car & truck & bus & train & mbike & bike & mIoU \\
		\hline
		PSPNet~\cite{pspnet} & 98.7 & 86.9 & 93.5 & 58.4 & 63.7 & 67.7 & 76.1 & 80.5 & 93.6 & 72.2 & 95.3 & 86.8 & 71.9 & 96.2 & 77.7 & 91.5 & 83.6 & 70.8 & 77.5 & 81.2 \\
		\hline
		ResNet38~\cite{resnet38}& 98.7 & 86.9 & 93.3 & 60.4 & 62.9 & 67.6 & 75.0 & 78.7 & 93.7 & 73.7 & 95.5 & 86.8 & 71.1 & 96.1 & 75.2 & 87.6 & 81.9 & 69.8 & 76.7 & 80.6 \\
		\hline
		InPlaceABN~\cite{inplaceabn}& 98.4 & 85.0 & 93.6 & 61.7 & 63.9 & 67.7 & 77.4 & 80.8 & 93.7 & 71.9 & 95.6 & 86.7 & 72.8 & 95.7 & 79.9 & \textbf{93.1} & \textbf{89.7} & 72.6 & 78.2 & 82.0 \\
		\hline
		DeepLabV3+~\cite{deeplabv3p}& 98.7 & 87.0 & 93.9 & 59.5 & 63.7 & 71.4 & 78.2 & 82.2 & 94.0 & 73.0 & 95.8 & 88.0 & 73.0 & 96.4 & 78.0 & 90.9 & 83.9 & 73.8 & 78.9 & 82.1 \\
		\hline
		Auto-Deeplab~\cite{auto-deeplab}& \textbf{98.8} & \textbf{87.6} & 93.8 & 61.4 & 64.4 & 71.2 & 77.6 & 80.9 & 94.1 & 72.7 & 96.0 & 87.8 & 72.8 & 96.5 & 78.2 & 90.9 & 88.4 & 69.0 & 77.6 & 82.1 \\
		\hline 
		DPC~\cite{DPC}& 98.7 & 87.1 & 93.8 & 57.7 & 63.5 & 71.0 & 78.0 & 82.1 & 94.0 & 73.3 & 95.4 & 88.2 & 74.5 & 96.5 & \textbf{81.2} & 93.3 & 89.0 & \textbf{74.1} & 79.0 & 82.6 \\
		\hline
		DRN-CRL~\cite{DRN} & \textbf{98.8} & 87.7 & 94.0 & 65.1 & 64.2 & 70.1 & 77.4 & 81.6 & 93.9 & \textbf{73.5} & 95.8 & 88.0 & 74.9 & 96.5 & 80.8 & 92.1 & 88.5 & 72.1 & 78.8 & 82.8 \\ 
		\hline
		Ours(ResNet101) & \textbf{98.8} & 87.5 & 94.0 & \textbf{65.3} & 66.4 & 71.0 & 77.6 & 81.0 & 94.0 & 72.6 & 95.9 & 87.6 & 75.0 & 96.3 & 80.2 & 90.3 & 87.9 & 72.9 & 78.9 & \textbf{82.9} \\ 
		\hline
		\hline
		Ours+ Mapillary & \textbf{98.8} & \textbf{87.7} & \textbf{94.2} & 65.0 & \textbf{66.7} & \textbf{73.1} & \textbf{79.3} &\textbf{82.4} & \textbf{94.2} & 72.9 & \textbf{96.0} & \textbf{88.4} & \textbf{76.2} & \textbf{96.5} & 79.8 & 89.6 & 87.7 & 74.0 & \textbf{80.0} & \textbf{83.3} \\
		\hline
	\end{tabular}
	\caption{Per-category results on Cityscapes test set trained with coarse data and Mapillary. Our model achieves the state of art results comparing with other methods using more stronger backbone.　Our method achieves better results than those use stronger backbone~\cite{DRN,inplaceabn}.
	}
	\label{tab:cityscapes_results_detail_coarse}
\end{table*}        

\subsection{Detailed Results on VOC2007}
Here we give the detailed detection results on VOC2007 shown in  and compared the results with previous detection methods, ours model achieves considerable results.

\subsection{More Visible Results on Cityscapes and COCO}
\begin{figure*}
	\centering
	\includegraphics[width=0.99\linewidth]{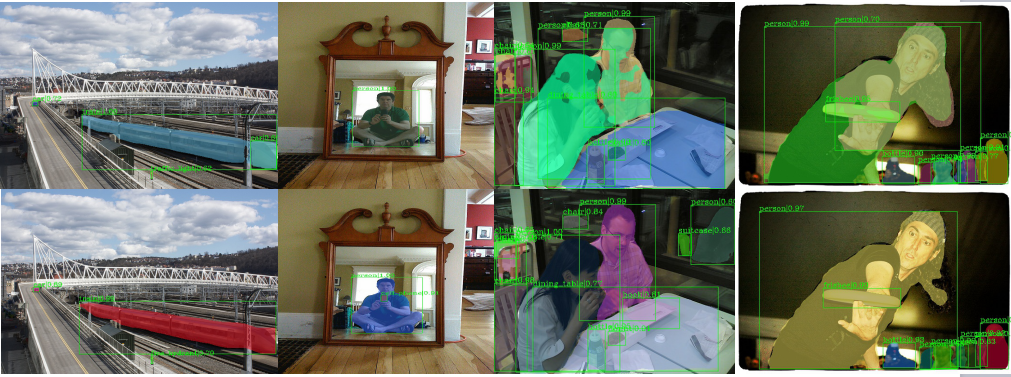}
	\caption{More detection and Segmentation Results on COCO
		First row: Mask-RCNN; Second row: + GALD
	}
	\label{fig:more_coco}
\end{figure*}

\begin{figure*}
	\centering
	\includegraphics[width=0.99\linewidth]{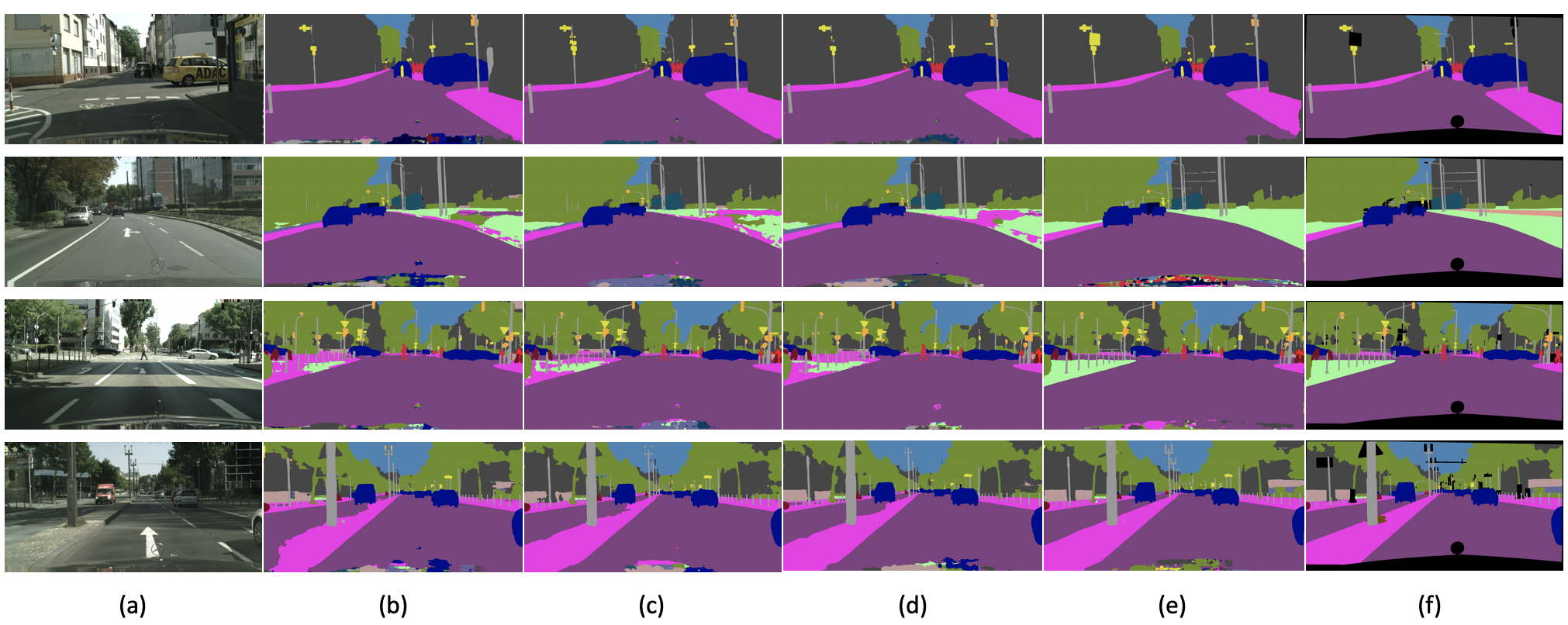}
	\caption{More cityscapes results: (a),input (b),FCN-res50 (c),+LD (d),+GA (e), +GALD (f),ground truth
	}
	\label{fig:more_city}
\end{figure*}

Here we show more results on Cityscapes and COCO dataset. 
COCO results are shown in Figure ~\ref{fig:more_coco}. 
Cityscapes results are shown in Figure ~\ref{fig:more_city}

\begin{table*}[t]
	\tiny
	\centering
	\setlength{\tabcolsep}{1.0pt}
	\begin{tabular}{ l | c c c c c c c c c c c c c c c c c c c c | c}
		\hline
		Method & aero & bike & bird & boat & bottle & bus & car & cat & chair & cow & table & dog & house& mbike & person & plant & sheep & sofa & train & tv & mAP(\%) \\
		\hline
		R-FCN\cite{R-FCN} & 79.9 & 87.2 & 81.5& 72.0& 69.8& 86.8& 88.5& 89.8& 67.0& \textbf{88.1}& 74.5& 89.8& \textbf{90.6}& 79.9& 81.2& 53.7& 81.8& 81.5& 85.9& 79.9& 80.5 \\ 
		\hline
		DSSD\cite{dssd} & 86.6& 86.2& 82.6& \textbf{74.9}& 62.5& \textbf{89.0}& 88.7& 88.8& 65.2& 87.0 & 78.7& 88.2& 89.0& 87.5& 83.7& 51.1& 86.3& \textbf{81.6}& 85.7& \textbf{83.7} & 81.5\\ 
		\hline
		DFPR\cite{dfpr} &  \textbf{92.0}& \textbf{88.2}& 81.1& 71.2& 65.7& 88.2& 87.9& \textbf{92.2}& 65.8& 86.5& \textbf{79.4}& \textbf{90.3}& 90.4& \textbf{89.3}& \textbf{88.6}& 59.4& \textbf{88.4}& 75.3& \textbf{89.2}& 78.5& 82.4\\ 
		\hline
		Faste-RCNN (base)\cite{faster-rcnn} & 86.5& 85.9& 82.9& 70.4& 70.4& 83.3& 88.1& 88.6& 66.0& 82.5& 74.6& 89.1& 87.1& 83.4& 85.8& 58.5& 84.8& 79.2& 85.9& 77.5 & 80.7\\ 
		\hline
		\hline    
		Ours & 86.7 & 87.7 & \textbf{85.3} & 74.5 & \textbf{74.4} & 86.1 & \textbf{89.0} & 89.5 & \textbf{71.2} & 87.5 & 77.5 & 89.0 & 87.9 & 85.4 & 86.5 & \textbf{59.8} & 86.1 & 80.8 & 87.6 & 83.2 & \textbf{83.0} \\ 
		\hline
	\end{tabular}
	\caption{PASCAL VOC 2007 test detection results. All models are trained with 07+12
		(07 trainval + 12 trainval). All the models are using ResNet101 as backbone.
	}
	\label{tab:VOC2007}
\end{table*}

\bibliography{egbib}

\begin{thebibliography}{45}
\providecommand{\natexlab}[1]{#1}
\providecommand{\url}[1]{\texttt{#1}}
\expandafter\ifx\csname urlstyle\endcsname\relax
  \providecommand{\doi}[1]{doi: #1}\else
  \providecommand{\doi}{doi: \begingroup \urlstyle{rm}\Url}\fi

\bibitem[Chen et~al.(2018{\natexlab{a}})Chen, Pang, Wang, Xiong, Li, Sun, Feng,
  Liu, Shi, Ouyang, Loy, and Lin]{mmdetection2018}
Kai Chen, Jiangmiao Pang, Jiaqi Wang, Yu~Xiong, Xiaoxiao Li, Shuyang Sun,
  Wansen Feng, Ziwei Liu, Jianping Shi, Wanli Ouyang, Chen~Change Loy, and
  Dahua Lin.
\newblock mmdetection.
\newblock \url{https://github.com/open-mmlab/mmdetection}, 2018{\natexlab{a}}.

\bibitem[{Chen} et~al.(2015){Chen}, {Papandreou}, {Kokkinos}, {Murphy}, and
  {Yuille}]{deeplabv1}
Liang-Chieh {Chen}, George {Papandreou}, Iasonas {Kokkinos}, Kevin {Murphy},
  and Alan~L. {Yuille}.
\newblock Semantic image segmentation with deep convolutional nets and fully
  connected {CRF}s.
\newblock \emph{ICLR}, 2015.

\bibitem[{Chen} et~al.(2017){Chen}, {Papandreou}, {Schroff}, and
  {Adam}]{deeplabv3}
Liang-Chieh {Chen}, George {Papandreou}, Florian {Schroff}, and Hartwig {Adam}.
\newblock Rethinking atrous convolution for semantic image segmentation.
\newblock \emph{arXiv preprint arXiv:1706.05587}, 2017.

\bibitem[Chen et~al.(2018{\natexlab{b}})Chen, Collins, Zhu, Papandreou, Zoph,
  Schroff, Adam, and Shlens]{DPC}
Liang-Chieh Chen, Maxwell Collins, Yukun Zhu, George Papandreou, Barret Zoph,
  Florian Schroff, Hartwig Adam, and Jon Shlens.
\newblock Searching for efficient multi-scale architectures for dense image
  prediction.
\newblock In \emph{NIPS}. 2018{\natexlab{b}}.

\bibitem[{Chen} et~al.(2018){Chen}, {Papandreou}, {Kokkinos}, {Murphy}, and
  {Yuille}]{deeplabv2}
Liang-Chieh {Chen}, George {Papandreou}, Iasonas {Kokkinos}, Kevin {Murphy},
  and Alan~L. {Yuille}.
\newblock Deeplab: Semantic image segmentation with deep convolutional nets,
  atrous convolution, and fully connected crfs.
\newblock \emph{PAMI}, 2018.

\bibitem[Chen et~al.(2018{\natexlab{a}})Chen, Zhu, Papandreou, Schroff, and
  Adam]{deeplabv3p}
Liang-Chieh Chen, Yukun Zhu, George Papandreou, Florian Schroff, and Hartwig
  Adam.
\newblock Encoder-decoder with atrous separable convolution for semantic image
  segmentation.
\newblock In \emph{ECCV}, 2018{\natexlab{a}}.

\bibitem[Chen et~al.(2018{\natexlab{b}})Chen, Kalantidis, Li, Yan, and
  Feng]{a2net}
Yunpeng Chen, Yannis Kalantidis, Jianshu Li, Shuicheng Yan, and Jiashi Feng.
\newblock A\^{}2-nets: Double attention networks.
\newblock In \emph{NIPS}. 2018{\natexlab{b}}.

\bibitem[Chen et~al.(2018{\natexlab{c}})Chen, Rohrbach, Yan, Yan, Feng, and
  Kalantidis]{graph_reason}
Yunpeng Chen, Marcus Rohrbach, Zhicheng Yan, Shuicheng Yan, Jiashi Feng, and
  Yannis Kalantidis.
\newblock Graph-based global reasoning networks.
\newblock \emph{arXiv preprint arXiv:1811.12814}, 2018{\natexlab{c}}.

\bibitem[Cordts et~al.(2016)Cordts, Omran, Ramos, Rehfeld, Enzweiler, Benenson,
  Franke, Roth, and Schiele]{Cityscapes}
Marius Cordts, Mohamed Omran, Sebastian Ramos, Timo Rehfeld, Markus Enzweiler,
  Rodrigo Benenson, Uwe Franke, Stefan Roth, and Bernt Schiele.
\newblock The cityscapes dataset for semantic urban scene understanding.
\newblock In \emph{CVPR}, 2016.

\bibitem[Dai et~al.(2016)Dai, Li, He, and Sun]{R-FCN}
Jifeng Dai, Yi~Li, Kaiming He, and Jian Sun.
\newblock R-fcn: Object detection via region-based fully convolutional
  networks.
\newblock In \emph{Advances in neural information processing systems}, pages
  379--387, 2016.

\bibitem[Everingham et~al.(2010)Everingham, Van~Gool, Williams, Winn, and
  Zisserman]{VOC}
Mark Everingham, Luc Van~Gool, Christopher~KI Williams, John Winn, and Andrew
  Zisserman.
\newblock The pascal visual object classes (voc) challenge.
\newblock \emph{IJCV}, 2010.

\bibitem[Falong~Shen and Zeng(2017)]{segmodel}
Shuicheng~Yan Falong~Shen, Gan~Rui and Gang Zeng.
\newblock Semantic segmentation via structured patch prediction, context crf
  and guidance crf.
\newblock In \emph{CVPR}, 2017.

\bibitem[Fu et~al.(2017)Fu, Liu, Ranga, Tyagi, and Berg]{dssd}
Cheng-Yang Fu, Wei Liu, Ananth Ranga, Ambrish Tyagi, and Alexander~C Berg.
\newblock Dssd: Deconvolutional single shot detector.
\newblock \emph{arXiv preprint arXiv:1701.06659}, 2017.

\bibitem[Fu et~al.(2018)Fu, Liu, Tian, Fang, and Lu]{DAnet}
Jun Fu, Jing Liu, Haijie Tian, Zhiwei Fang, and Hanqing Lu.
\newblock Dual attention network for scene segmentation.
\newblock \emph{arXiv preprint arXiv:1809.02983}, 2018.

\bibitem[Huang et~al.(2018)Huang, Wang, Huang, Huang, Wei, and Liu]{ccnet}
Zilong Huang, Xinggang Wang, Lichao Huang, Chang Huang, Yunchao Wei, and Wenyu
  Liu.
\newblock Ccnet: Criss-cross attention for semantic segmentation.
\newblock \emph{arXiv preprint arXiv:1811.11721}, 2018.

\bibitem[Ke et~al.(2018)Ke, Hwang, Liu, and Yu]{aaf}
Tsung-Wei Ke, Jyh-Jing Hwang, Ziwei Liu, and Stella~X. Yu.
\newblock Adaptive affinity fields for semantic segmentation.
\newblock In \emph{ECCV}, 2018.

\bibitem[Kong et~al.(2018)Kong, Sun, Tan, Liu, and Huang]{dfpr}
Tao Kong, Fuchun Sun, Chuanqi Tan, Huaping Liu, and Wenbing Huang.
\newblock Deep feature pyramid reconfiguration for object detection.
\newblock In \emph{Proceedings of the European Conference on Computer Vision
  (ECCV)}, pages 169--185, 2018.

\bibitem[Li and Gupta(2018)]{beyond_grids}
Yin Li and Abhinav Gupta.
\newblock Beyond grids: Learning graph representations for visual recognition.
\newblock In \emph{NIPS}. 2018.

\bibitem[Lin et~al.(2014)Lin, Maire, Belongie, Hays, Perona, Ramanan,
  Doll{\'a}r, and Zitnick]{COCO_dataset}
Tsung-Yi Lin, Michael Maire, Serge Belongie, James Hays, Pietro Perona, Deva
  Ramanan, Piotr Doll{\'a}r, and C~Lawrence Zitnick.
\newblock Microsoft coco: Common objects in context.
\newblock In \emph{ECCV}. Springer, 2014.

\bibitem[{Lin} et~al.(2017){Lin}, {Dollár}, {Girshick}, {He}, {Hariharan}, and
  {Belongie}]{fpn}
Tsung-Yi {Lin}, Piotr {Dollár}, Ross~B. {Girshick}, Kaiming {He}, Bharath
  {Hariharan}, and Serge~J. {Belongie}.
\newblock Feature pyramid networks for object detection.
\newblock In \emph{CVPR}, 2017.

\bibitem[Liu et~al.(2019)Liu, Chen, Schroff, Adam, Hua, Yuille, and
  Fei-Fei]{auto-deeplab}
Chenxi Liu, Liang-Chieh Chen, Florian Schroff, Hartwig Adam, Wei Hua, Alan
  Yuille, and Li~Fei-Fei.
\newblock Auto-deeplab: Hierarchical neural architecture search for semantic
  image segmentation.
\newblock \emph{arXiv preprint arXiv:1901.02985}, 2019.

\bibitem[Liu et~al.(2015)Liu, Rabinovich, and Berg]{parsenet}
Wei Liu, Andrew Rabinovich, and Alexander~C Berg.
\newblock Parsenet: Looking wider to see better.
\newblock \emph{arXiv preprint arXiv:1506.04579}, 2015.

\bibitem[{Long} et~al.(2015){Long}, {Shelhamer}, and {Darrell}]{fcn}
Jonathan {Long}, Evan {Shelhamer}, and Trevor {Darrell}.
\newblock Fully convolutional networks for semantic segmentation.
\newblock In \emph{CVPR}, 2015.

\bibitem[Luo et~al.(2016)Luo, Li, Urtasun, and Zemel]{luo2016understanding}
Wenjie Luo, Yujia Li, Raquel Urtasun, and Richard Zemel.
\newblock Understanding the effective receptive field in deep convolutional
  neural networks.
\newblock In \emph{Advances in neural information processing systems}, pages
  4898--4906, 2016.

\bibitem[Neuhold et~al.(2017)Neuhold, Ollmann, Rota~Bulo, and
  Kontschieder]{mapillary}
Gerhard Neuhold, Tobias Ollmann, Samuel Rota~Bulo, and Peter Kontschieder.
\newblock The mapillary vistas dataset for semantic understanding of street
  scenes.
\newblock In \emph{ICCV}, 2017.

\bibitem[Ren et~al.(2015)Ren, He, Girshick, and Sun]{faster-rcnn}
Shaoqing Ren, Kaiming He, Ross Girshick, and Jian Sun.
\newblock Faster r-cnn: Towards real-time object detection with region proposal
  networks.
\newblock In \emph{Advances in neural information processing systems}, pages
  91--99, 2015.

\bibitem[Rota~Bul{\`o} et~al.(2018)Rota~Bul{\`o}, Porzi, and
  Kontschieder]{inplaceabn}
Samuel Rota~Bul{\`o}, Lorenzo Porzi, and Peter Kontschieder.
\newblock In-place activated batchnorm for memory-optimized training of dnns.
\newblock In \emph{Proceedings of the IEEE Conference on Computer Vision and
  Pattern Recognition}, pages 5639--5647, 2018.

\bibitem[Wang et~al.(2018)Wang, Girshick, Gupta, and He]{Nonlocal}
Xiaolong Wang, Ross Girshick, Abhinav Gupta, and Kaiming He.
\newblock Non-local neural networks.
\newblock In \emph{CVPR}, 2018.

\bibitem[Woo et~al.(2018)Woo, Park, Lee, and So~Kweon]{cbam}
Sanghyun Woo, Jongchan Park, Joon-Young Lee, and In~So~Kweon.
\newblock Cbam: Convolutional block attention module.
\newblock In \emph{The European Conference on Computer Vision (ECCV)},
  September 2018.

\bibitem[Wu et~al.(2016)Wu, Shen, and Hengel]{SegOHEM}
Zifeng Wu, Chunhua Shen, and Anton van~den Hengel.
\newblock High-performance semantic segmentation using very deep fully
  convolutional networks.
\newblock \emph{arXiv preprint arXiv:1604.04339}, 2016.

\bibitem[{Wu} et~al.(2016){Wu}, {Shen}, and van~den {Hengel}]{resnet38}
Zifeng {Wu}, Chunhua {Shen}, and Anton van~den {Hengel}.
\newblock Wider or deeper: Revisiting the resnet model for visual recognition.
\newblock \emph{arXiv preprint arXiv:1611.10080}, 2016.

\bibitem[Yang et~al.(2018)Yang, Yu, Zhang, Li, and Yang]{denseaspp}
Maoke Yang, Kun Yu, Chi Zhang, Zhiwei Li, and Kuiyuan Yang.
\newblock Denseaspp for semantic segmentation in street scenes.
\newblock In \emph{CVPR}, 2018.

\bibitem[Yu et~al.(2018{\natexlab{a}})Yu, Wang, Peng, Gao, Yu, and
  Sang]{bisenet}
Changqian Yu, Jingbo Wang, Chao Peng, Changxin Gao, Gang Yu, and Nong Sang.
\newblock Bisenet: Bilateral segmentation network for real-time semantic
  segmentation.
\newblock In \emph{ECCV}, 2018{\natexlab{a}}.

\bibitem[Yu et~al.(2018{\natexlab{b}})Yu, Wang, Peng, Gao, Yu, and Sang]{dfn}
Changqian Yu, Jingbo Wang, Chao Peng, Changxin Gao, Gang Yu, and Nong Sang.
\newblock Learning a discriminative feature network for semantic segmentation.
\newblock In \emph{CVPR}, 2018{\natexlab{b}}.

\bibitem[Yuan and Wang(2018)]{ocnet}
Yuhui Yuan and Jingdong Wang.
\newblock Ocnet: Object context network for scene parsing.
\newblock \emph{arXiv preprint arXiv:1809.00916}, 2018.

\bibitem[Yue et~al.()Yue, Sun, Yuan, Zhou, Ding, and Xu]{cgnl}
Kaiyu Yue, Ming Sun, Yuchen Yuan, Feng Zhou, Errui Ding, and Fuxin Xu.
\newblock Compact generalized non-local network.
\newblock In \emph{NIPS}.

\bibitem[Zhang et~al.(2018)Zhang, Dana, Shi, Zhang, Wang, Tyagi, and
  Agrawal]{encodingnet}
Hang Zhang, Kristin Dana, Jianping Shi, Zhongyue Zhang, Xiaogang Wang, Ambrish
  Tyagi, and Amit Agrawal.
\newblock Context encoding for semantic segmentation.
\newblock In \emph{CVPR}, June 2018.

\bibitem[Zhang et~al.(2019{\natexlab{a}})Zhang, Zhang, Wang, and
  Xie]{CoCurrentNet}
Hang Zhang, Han Zhang, Chenguang Wang, and Junyuan Xie.
\newblock Co-occurrent features in semantic segmentation.
\newblock In \emph{CVPR}, June 2019{\natexlab{a}}.

\bibitem[Zhang et~al.(2019{\natexlab{b}})Zhang, Li, Arnab, Yang, Tong, and
  Torr]{zhangli_dgcn}
Li~Zhang, Xiangtai Li, Anurag Arnab, Kuiyuan Yang, Yunhai Tong, and Philip~HS
  Torr.
\newblock Dual graph convolutional network for semantic segmentation.
\newblock In \emph{British Machine Vision Conference}, 2019{\natexlab{b}}.

\bibitem[Zhang et~al.(2019{\natexlab{c}})Zhang, Xu, Arnab, and
  Torr]{zhang2019dynamic}
Li~Zhang, Dan Xu, Anurag Arnab, and Philip~HS Torr.
\newblock Dynamic graph message passing network.
\newblock \emph{arXiv preprint arXiv:1908.06959}, 2019{\natexlab{c}}.

\bibitem[Zhang et~al.(2017)Zhang, Tang, Zhang, Li, and Yan]{sac}
Rui Zhang, Sheng Tang, Yongdong Zhang, Jintao Li, and Shuicheng Yan.
\newblock Scale-adaptive convolutions for scene parsing.
\newblock In \emph{ICCV}, 2017.

\bibitem[{Zhao} et~al.(2017){Zhao}, {Shi}, {Qi}, {Wang}, and {Jia}]{pspnet}
Hengshuang {Zhao}, Jianping {Shi}, Xiaojuan {Qi}, Xiaogang {Wang}, and Jiaya
  {Jia}.
\newblock Pyramid scene parsing network.
\newblock In \emph{CVPR}, 2017.

\bibitem[Zhao et~al.(2018)Zhao, Zhang, Liu, Shi, Change~Loy, Lin, and
  Jia]{psanet}
Hengshuang Zhao, Yi~Zhang, Shu Liu, Jianping Shi, Chen Change~Loy, Dahua Lin,
  and Jiaya Jia.
\newblock Psanet: Point-wise spatial attention network for scene parsing.
\newblock In \emph{ECCV}, 2018.

\bibitem[Zhou et~al.(2014)Zhou, Khosla, Lapedriza, Oliva, and
  Torralba]{zhou2014object}
Bolei Zhou, Aditya Khosla, Agata Lapedriza, Aude Oliva, and Antonio Torralba.
\newblock Object detectors emerge in deep scene cnns.
\newblock \emph{arXiv preprint arXiv:1412.6856}, 2014.

\bibitem[Zhuang et~al.(2018)Zhuang, Yang, Tao, Ma, Zhang, Li, Jia, Xie, and
  Gao]{DRN}
Yueqing Zhuang, Fan Yang, Li~Tao, Cong Ma, Ziwei Zhang, Yuan Li, Huizhu Jia,
  Xiaodong Xie, and Wen Gao.
\newblock Dense relation network: Learning consistent and context-aware
  representation for semantic image segmentation.
\newblock In \emph{ICIP}. IEEE, 2018.

\end{thebibliography}
\end{document}